\documentclass[11pt,a4paper]{article}
\usepackage[hyperref]{emnlp-ijcnlp-2019}
\usepackage{hyperref}
\usepackage{url}
\newcommand{\ignore}[1]{}

\newcommand{\hotpot}{\textsc{HotpotQA}\xspace}

\newcommand{\whop}{\textsc{Wikihop}\xspace}
\newcommand{\bert}{\textsc{Bert}\xspace}
\newcommand{\bm}{\textsc{BM25}\xspace}
\newcommand{\cls}{\textsc{[cls]}\xspace}
\newcommand{\prf}{\textsc{Prf}\xspace}

\usepackage{pslatex}
\usepackage{latexsym}
\usepackage{amsfonts}
\usepackage{mathtools}
\usepackage{graphicx}
\usepackage{amsmath}
\usepackage{xspace}
\usepackage{tikz-dependency}
\usepackage{balance}
\usepackage{enumitem}
\usepackage{mathtools}
\usepackage{algorithm}
\usepackage{algorithmicx}
\usepackage{eqparbox}

\usepackage{algcompatible}

\usepackage{footnote}
\makesavenoteenv{tabular}

\usepackage{varwidth}
\usepackage{pifont}
\usepackage{float}
\usepackage{framed} 
\usepackage{verbatim}
\usepackage[rightcaption]{sidecap}
\usepackage{subfig}
\usepackage{wrapfig}
\usepackage{multirow}
\DeclareSymbolFont{extraup}{U}{zavm}{m}{n}
\DeclareMathSymbol{\varheartsuit}{\mathalpha}{extraup}{86}

\newcommand\blfootnote[1]{%
  \begingroup
  \renewcommand\thefootnote{}\footnote{#1}%
  \addtocounter{footnote}{-1}%
  \endgroup
}
\usepackage{xargs} 
\usepackage[colorinlistoftodos,prependcaption,textsize=tiny]{todonotes}

\usepackage{marginnote}
\usepackage{lipsum}

\usepackage{microtype}
\usepackage{booktabs}
\usepackage{rotating}

\aclfinalcopy 

    \title{Multi-step Entity-centric Information Retrieval for \\Multi-Hop Question Answering}

\author{Ameya Godbole$^{1*}$, Dilip Kavarthapu$^{1*}$, Rajarshi Das$^{1*}$, \\\textbf{Zhiyu Gong$^{1}$, Abhishek Singhal$^{1}$, Hamed Zamani$^{1}$}, \\\textbf{Mo Yu$^{2}$, Tian Gao$^{2}$, Xiaoxiao Guo$^{2}$, Manzil Zaheer$^{3}$ and Andrew McCallum$^{1}$}\\
$^1$University of Massachusetts Amherst, \\$^2$IBM Research, $^3$Google Research
}

\date{}

\begin{document}
\maketitle
\begin{abstract}
\blfootnote{$^\ast$ Equal contribution. Correspondence to \{agodbole, rajarshi\}@cs.umass.edu}Multi-hop question answering (QA) requires an information retrieval (IR) system that can find \emph{multiple} supporting evidence needed to answer the question, making the retrieval process very challenging. This paper introduces an IR technique that uses information of entities present in the initially retrieved evidence to learn to `\emph{hop}' to other relevant evidence. In a setting, with more than \textbf{5 million} Wikipedia paragraphs, our approach leads to significant boost in retrieval performance. The retrieved evidence also increased the performance of an existing QA model (without any training) on the \hotpot benchmark by \textbf{10.59} F1.
\end{abstract}

\section{Introduction}

\begin{figure}
    \centering
    \vspace{-2mm}
    \includegraphics[width=\columnwidth]{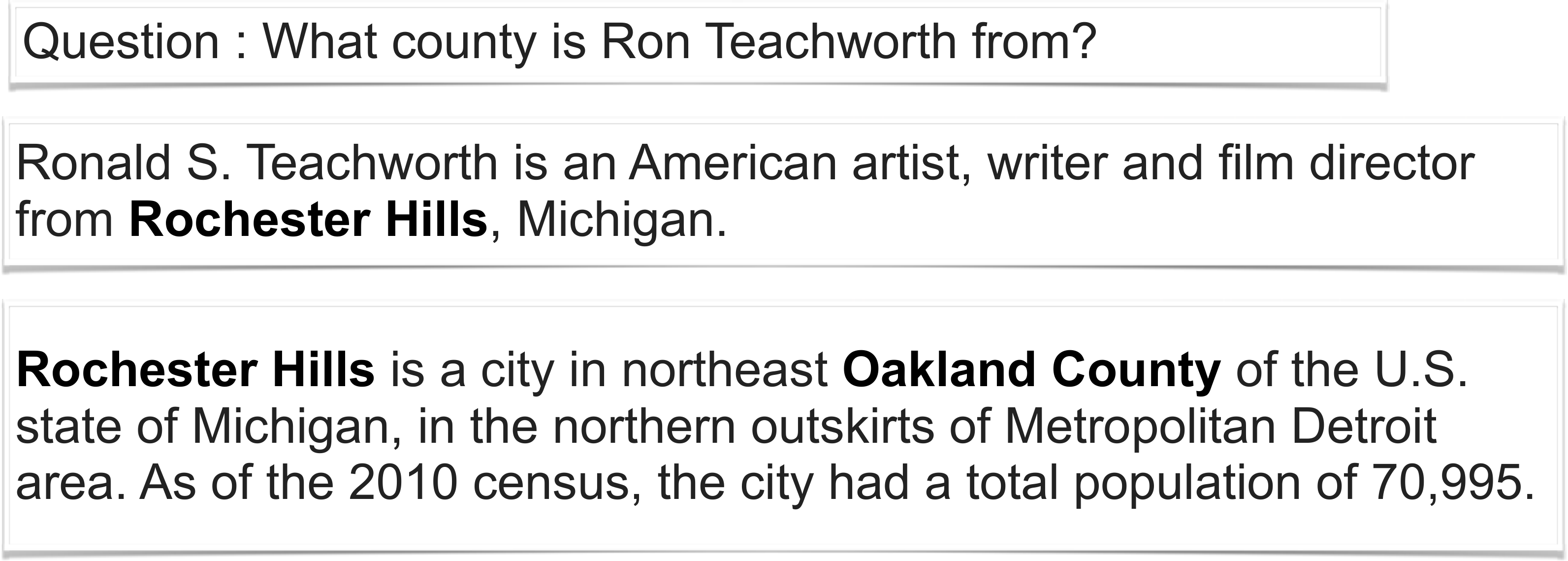}
    \caption{Multi-hop questions require finding multiple evidence and the target document containing the answer has very little lexical overlap with the question.}
    \label{fig:multi_hop_ex1}
    \vspace{-3mm}
\end{figure}

Multi-hop QA requires finding multiple supporting evidence, and reasoning over them in order to answer a question \cite{welbl2018constructing,talmor2018web,yang2018hotpotqa}. For example, to answer the question shown in figure~\ref{fig:multi_hop_ex1}, the QA system has to retrieve two different paragraphs and reason over them. Moreover, the paragraph containing the answer to the question has very little lexical overlap with the question, making it difficult for search engines to retrieve them from a large corpus. For instance, the accuracy of a \bm retriever for finding \emph{all} supporting evidence for a question decreases from 53.7\% to 25.9\% on the `easy' and `hard' subsets of the \hotpot training dataset.\footnote{According to \citet{yang2018hotpotqa}, the easy (hard) subset primarily requires single (multi) hop reasoning. We only consider queries that have answers as spans in at least one paragraph.}

We hypothesize that an effective retriever for multi-hop QA should have the ``\emph{hopiness}'' built into it, by design. That is, after retrieving an initial set of documents, the retriever should be able to ``hop'' onto other documents, if required. We note that, many supporting evidence often share common (\emph{bridge}) entities between them (e.g. ``Rochester Hills'' in figure~\ref{fig:multi_hop_ex1}). In this work, we introduce a model that uses information about entities present in an initially retrieved paragraph to jointly find a passage of text \emph{describing} the entity (\emph{entity-linking}) and also determining whether that passage would be relevant to answer the multi-hop query.


A major component of our retriever is a re-ranker model that uses contextualized entity representation obtained from a pre-trained \bert \cite{devlin2018bert} language model. Specifically, the entity representation is obtained by feeding the query and a Wikipedia paragraph describing the entity to a \bert model. The re-ranker uses representation of both the initial paragraph and the representation of all the entities within it to determine which evidence to gather next. 


\begin{figure*}
\centering
    \includegraphics[width=1.75\columnwidth]{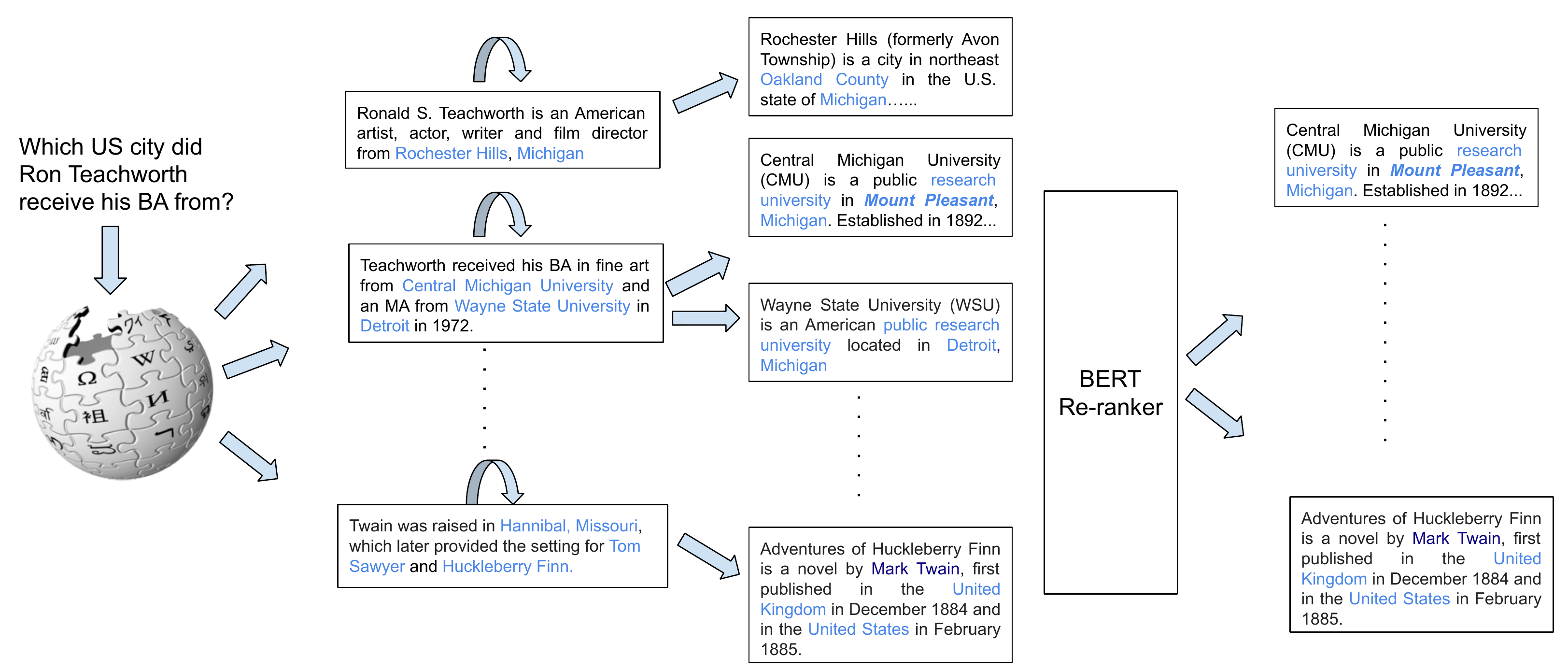}
\caption{Overview of our approach. We use the entity mentions present in the initially retrieved paragraphs to link to paragraphs describing them. Next, the \bert-based re-ranker scores the chain of initial and the entity-describing paragraph. Note the presence of self-loop from the initial paragraphs to accommodate for questions that do not require `hopping' to a new paragraph.}
\label{fig:model}
\end{figure*}
Essentially, our method introduces a new way of \emph{multi-step} retrieval that uses information about intermediate entities. 
A standard way of doing multi-step retrieval is via \emph{pseudo-relevance feedback} \cite{xu2017quary,lavrenko2017relevance} in which relevant terms from initial retrieved documents are used to reformulate the initial question. A few recent works learn to reformulate the query using task specific reward such as document recall or performance on a QA task \cite{nogueira2017task,buck2017ask,das2018multi}. However, these methods do not necessarily use the information about entities present in the evidence as they might not be the more frequent/salient terms in it. Empiricially, our method outperforms all of these methods significantly for multi-hop QA. Our work is most closely related to the recently proposed \bert re-ranker model of \citet{nogueira2019passage}. However, unlike us, they do not model the chains of evidence paragraphs required for a multi-hop question. Secondly, they also do not have a entity linking component to identify the relevant paragraphs. Our model out-performs them for multi-hop QA.


To summarize, this paper presents an entity-centric IR approach that jointly performs entity linking and effectively finds relevant evidence required for questions that need multi-hop reasoning from a large corpus containing millions of paragraphs. When the retrieved paragraphs are supplied to the baseline QA model introduced in \citet{yang2018hotpotqa}, it improved the QA performance on the hidden test set by 10.59 F1 points.\footnote{Code, pre-trained models and retrieved paragraphs are released --- \url{https://github.com/ameyagodbole/entity-centric-ir-for-multihop-qa}}
\section{Methodology}
\label{sec:model}
Our approach is summarized in Figure~\ref{fig:model}. The first component of our model is a standard IR system that takes in a natural language query `$\mathrm{Q}$' and returns an initial set of evidence. For our experiments, we use the popular \bm retriever, but this component can be replaced by any IR model. We assume that all spans of entity mentions have been identified in the paragraph text by a one-time pre-processing, with an entity tagger.\footnote{We plan to explore joint learning of entity tagging with linking and retrieval as future work.}

\paragraph{Entity Linking} The next component of our model is an entity linker that finds an introductory Wikipedia paragraph describing the entity, corresponding to each entity mention. Several IR approaches \cite{xiong2016bag,raviv2016document} use an off-the-shelf entity linker. However, most entity linking systems \cite{ganea,raiman2018deeptype} have been trained on Wikipedia data and hence using an off-the-shelf linker would be unfair, since there exists a possibility of test-time leakage. To ensure strictness, we developed our own simple linking strategy. Following the standard approach of using mention text and hyper-link information \cite{cucerzan2007large,ji2011knowledge}, we create a mapping (alias table) between them. The alias table stores mappings between a mention string (e.g. ``Bill'') and various entities it can refer to (e.g. Bill Clinton, Billy Joel, etc).
The top-40 documents returned by the \bm retriever on the dev and test queries are also ignored while building the alias table.
At test time, our re-ranker considers all the candidate entity paragraphs that a mention is linked to via the alias table. Although simple, we find this strategy to work well for our task and we plan to use a learned entity linker for future work.

\paragraph{Re-ranker} The next component of our model is a \bert-based re-ranker that ranks the chains of paragraphs obtained from the previous two components of the model. Let $\mathrm{Q}$ denote the query, $\mathrm{D}$ denote a paragraph in the initial set of paragraphs returned by the \bm retriever. Let $\mathrm{e}$ denote an entity mention present in $\mathrm{D}$ and $\mathrm{E}$ be the linked document returned by the linker for $\mathrm{e}$. If there are multiple linked documents, we consider all of them. Although our retriever is designed for multi-hop questions, in a general setting, most questions are not multi-hop in nature. Therefore to account for questions that do not need hopping to a new paragraph, we also add a `self-link' (Figure~\ref{fig:model}) from each of the initial retrieved paragraph, giving the model the ability to stay in the same paragraph.

To train the re-ranker, we form \emph{query-dependent} passage representation for both $\mathrm{D}$ and $\mathrm{E}$. The query $\mathrm{Q}$ and the paragraph $\mathrm{E}$ are concatenated and fed as input to a \bert encoder and the corresponding \cls token forms the entity representation $\mathbf{e}$. Similarly, the document representation $\mathbf{d}$ is set to the embedding of the \cls token obtained after encoding the concatenation of $\mathrm{Q}$ and $\mathrm{D}$. The final score that the entity paragraph $\mathrm{E}$ is relevant to $\mathrm{Q}$ is computed by concatenating the two query-aware representation $\mathbf{d}$ and $\mathbf{e}$ and passing it through a 2-layer feed-forward network as before. It should be noted, the final score is determined by both the evidence paragraphs $\mathrm{D}$ and $\mathrm{E}$ and as we show empirically, not considering both leads to decrease in performance.

During training, we mark a chain of paragraphs as a positive example, if the last paragraph of the chain is present in the supporting facts, since that is a chain of reasoning that led to a relevant paragraph. All other paragraph chains are treated as negative examples. In our experiments, we consider chains of length 2, although extending to longer chains is straightforward. We use a simple binary cross-entropy loss to train the network.

\section{Experiments}
For all our experiment, unless specified otherwise, we use the open domain corpus\footnote{\url{https://hotpotqa.github.io/wiki-readme.html}} released by \citet{yang2018hotpotqa} which contains over 5.23 million Wikipedia abstracts (introductory paragraphs). To identify spans of entities, we use the implementation of the state-of-the-art entity tagger presented in \citet{peters2018deep}.\footnote{\url{https://allennlp.org/models}} For the \bert encoder, we use the \textsc{Bert-base-uncased} model.\footnote{\url{https://github.com/google-research/bert}} We use the implementation of widely-used \bm retrieval available in Lucene.\footnote{\url{https://lucene.apache.org/}} 


\subsection{IR for MultiHop QA}
We introduce a new way of doing multi-step retrieval. A popular way of doing it in traditional IR systems is via pseudo-relevance feedback (\textsc{Prf}). The \prf methods assume that the top retrieved documents in response to a given query are relevant. Based on this assumption, they expand the query in a weighted manner. \prf has been shown to be effective in various retrieval settings~\cite{xu2017quary}. We compare with two widely used \prf models  --- The Rocchio's algorithm on top of the TF-IDF retrieval model (\textsc{Prf-tfidf}) \cite{Rocchio:1971} and the relevance model (RM3) based on the language modeling framework in information retrieval (\textsc{Prf-rm}) \cite{lavrenko2017relevance}. Following prior work~\cite{nogueira2017task}, we use query likelihood retrieval model with Dirichlet prior smoothing \cite{zhai2017study} for first retrieval run. 

\citet{nogueira2017task} proposed a new way of query reformulation --- incorporating reward from a document-relevance task (\textsc{Prf-task}) and training using reinforcement learning. 
Recently, \citet{nogueira2019passage} proposed a \bert based passage re-ranker (\bert-re-ranker) that has achieved excellent performance in several IR benchmarks. 
But, its performance has not been evaluated on multi-hop queries till now. 
For a fair comparison with our model which looks at paragraphs corresponding to entities, we use top 200 paragraphs retrieved by the initial IR model for \bert-re-ranker instead of 25 for our model.\footnote{There were 2.725 entities in the paragraph on average}

\begin{table}
    \small
    \centering
    \begin{tabular}{@{} l c c c c c @{}}\toprule
    & \multicolumn{4}{c}{\textsc{Accuracy}  } & \\
    \cline{2-5}\noalign{\smallskip}
    Model & @2 & @5 & @10 & @20 & \textsc{map} \\\midrule
    \bm & 0.093 & 0.191 & 0.259&  0.324 & 0.412\\
    \textsc{Prf-tfidf} & 0.088 & 0.157 & 0.204  & 0.258 & 0.317\\
    \textsc{Prf-rm}  & 0.083 & 0.175 & 0.242 & 0.296 & 0.406\\ 
    \textsc{Prf-task}  & 0.097 & 0.198 & 0.267 & 0.330 & 0.420\\\midrule
    \bert-re-ranker & 0.146 & 0.271 & 0.347 & 0.409 & 0.470\\
    \textsc{Query+E-doc} & 0.101 & 0.223 & 0.301 & 0.367 & 0.568\\
    \midrule
    Our Model &\textbf{0.230} & \textbf{0.482}&  \textbf{0.612}& \textbf{0.674} & \textbf{0.654}\\\bottomrule
    \end{tabular}
    \caption{Retrieval performance of models on the \hotpot benchmark. A successful retrieval is when \emph{all} the relevant passages for a question are retrieved from more than 5 million paragraphs in the corpus.}
    \label{tab:retriever}
    \vspace{-3mm}
\end{table}

\begin{figure}
    \centering
    \vspace{-2mm}
    \includegraphics[width=0.7\columnwidth]{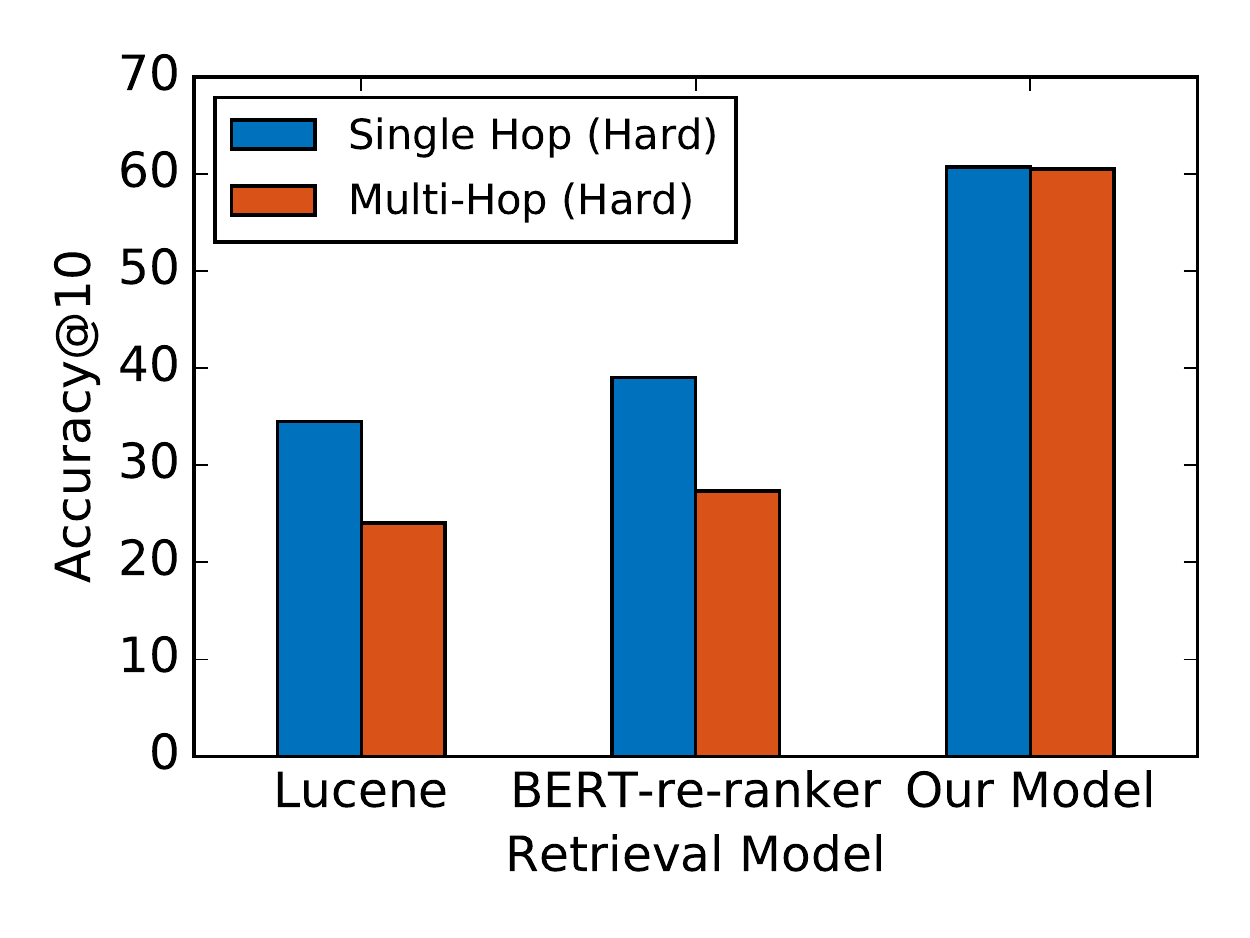}
    \vspace{-3mm}
    \caption{Our retrieval model works equally well for single-hop queries. This can be attributed to the presence of self-loops in the model which can make the model not hop to a different paragraph, if not required.}
    \label{fig:single_vs_multi}
    \vspace{-3mm}
\end{figure}

Table~\ref{tab:retriever} reports the accuracy(@$k$) of retrieving \emph{all} the relevant paragraphs required for answering a question in \hotpot.\footnote{Since, the supporting passage information is only present for train \& validation set, we consider the validation set as our hidden test set and consider a subset of train as validation set.} We compare the supporting passage annotation present in the dataset with the retrieved paragraph and retrieval is only correct when all the supporting paragraphs are in the top $k$ retrieved passages. We also report the mean average precision score (\textsc{map}) which is a strict metric that takes into account the relative position of the relevant document in the ranked list \cite{kadlec2017knowledge}. As we can see, our retrieval technique vastly outperforms other existing retrieval systems with an absolute increase of \textbf{26.5}\% (accuracy@10) and \textbf{18.4}\% (\textsc{map}), when compared to \bert-re-ranker. The standard \prf techniques do not perform well for this task. This is primarily because the \prf methods rely on statistical features like frequency of terms in the document, and fail to explicitly use information about entities, that may not be frequently occurring the paragraph. In fact, their performance is a little behind the standard retrieval results of \bm, suggesting that this benchmark dataset needs entity-centric information retrieval. The \textsc{Prf-task} does slightly better than other \prf models, showing that incorporating task-specific rewards can be beneficial. However, as we find, RL approaches are slow to converge\footnote{Training took $\sim$2 weeks for comparable performance.} as rewards from a down-stream tasks are sparse and action space in information retrieval is very large.\\
\textbf{Ablations}. We investigate whether modeling the chain of paragraphs needed to reach the final paragraph is important or not. As an ablation, we ignore the representation of the initial retrieved document $\mathrm{D}_1$ and only consider the final document representing the entity (\textsc{Query+E-doc}). Table~\ref{tab:retriever} shows that, indeed modeling the chain of documents is important. This makes intuitive sense, since to answer questions such as the county where a person is from (figure~\ref{fig:multi_hop_ex1}), modeling context about the person, should be helpful.
We also evaluate, if our model performs well on single-hop questions as well. This evaluation is a bit tricky to do in \hotpot, since the evaluataion set only contains questions from `hard' subset \cite{yang2018hotpotqa}. However, within that hard subset, we find the set of question, that has the answer span present in \emph{all} the supporting passages (\textsc{single-hop (hard)}) and only in \emph{one} of the supporting passages (\textsc{multi-hop (hard)})\footnote{There were 1184 \textsc{single-hop (hard)} and 4734 \textsc{multi-hop (hard)} queries.}. The intuition is that if there are multiple evidence containing the answer spans then it might be a little easier for a downstream QA model to identify the answer span. Figure~\ref{fig:single_vs_multi} shows that our model performs equally well on both type of queries and hence can be applied in a practical setting.
\subsection{Performance on \hotpot}
Table~\ref{tab:QA} shows the performance on the QA task. We were able to achieve better scores than reported in the baseline reader model of \citet{yang2018hotpotqa} by using Adam \cite{kingma2014adam} instead of standard SGD (our re-implementation). Next, we use the top-10 paragraphs retrieved by our system from the entire corpus and feed it to the reader model. We achieve a \textbf{10.59} absolute increase in F1 score than the baseline. It should be noted that we use the simple baseline reader model and we are confident that we can achieve better scores by using more sophisticated reader architectures, e.g. using \bert based architectures. 
Our results show that retrieval is an important component of an open-domain system and equal importance should be given to both the retriever and reader component.
\begin{table}
    \small
    \centering
    \begin{tabular}{@{} l c c @{}}\toprule
    Model & EM & F1   \\\midrule
    Baseline Reader \cite{yang2018hotpotqa} & 23.95 & 32.89\\
    Our re-implementation & 26.06 & 35.67 \\  
      + retrieved result &\textbf{35.36} & \textbf{46.26} \\\bottomrule
    \end{tabular}
    \caption{Performance on QA task on hidden test set of \hotpot after adding the retrieved paragraphs}
    \label{tab:QA}
    \vspace{-3mm}
\end{table}
\subsection{Zero-shot experiment on Wikihop}
We experiment if our model trained on \hotpot can generalize to another multi-hop dataset -- \whop \cite{welbl2018constructing}, without any training. In the \whop dataset, a set of candidate introductory Wikipedia paragraphs are given per question. Hence, we do not need to use our initial \bm retriever.

We assign the first entity mention occurring in a paragraph as the textual description of that entity. For instance, if the first entity mention in the paragraph is `Mumbai', we assign that paragraph as the textual description for the entity `Mumbai'. This assumption is often true for the introductory paragraphs of a Wikipedia article. Next, we perform entity linking of mentions by just simple string matching (i.e. linking strings such as `mumbai' to the previous paragraph). After constructing a small subgraph from the candidate paragraphs, we apply our model trained on \hotpot. The baseline models we compare to are a \bm retriever and a \bert-re-ranker model of \citep{nogueira2019passage} that ranks all the candidate supporting paragraphs for the question. Table~\ref{tab:wikihop} shows our model outperforms both models in zero-shot setting.

\begin{table}
    \small
    \centering
    \begin{tabular}{@{} l c c @{}}\toprule
    Model & acc@2 & acc@5   \\\midrule
    \bm & 0.06 & 0.30\\
    \bert-re-ranker (zs) & 0.08 & 0.27 \\  
    Our Model (zs) &\textbf{0.10} & \textbf{0.41} \\\bottomrule
    \end{tabular}
    \caption{Zero-shot (zs) IR results on \whop.}
    \label{tab:wikihop}
    \vspace{-3mm}
\end{table}

\section{Related Work}
\textbf{Document retrieval using entities}. Analysis of web-search query logs has revealed that there is a large portion of entity seeking queries \cite{liu2015latent}. There exists substantial work on modeling documents with entities occurring in them. For example, \citet{xiong2016bag} represents a document with bag-of-entities and \citet{raviv2016document} use entity-based language modeling for document retrieval. However, they depend on an off-the-shelf entity tagger, where as we jointly perform linking and retrieval. Moreover, we use contextualized entity representations using pre-trained LMs which have been proven to be better than bag-of-words approaches.
There has been a lot of work which leverages knowledge graphs (KGs) to learn better entity representations \cite{xiong2015esdrank,xiong2017word,liu2018entity} and for better query reformulation \cite{cao2008selecting,dalton2014entity,dietz2014umass}. Our work is not tied to any specific KG schema, instead we encode entities using its text description.  
\\
\textbf{Neural ranking} models have shown great potential and have been widely adopted in the IR community \cite[inter-alia]{dehghani2017nrm,guo:2019,mitra2017learning,zamani2018neural}. Bag-of-words and contextual embedding models, such as word2vec and BERT, have also been explored extensively for various IR tasks, from document to sentence-level retrieval~\cite{padigela2019,zamani2016emb,zamani2017relwe}.


\section{Conclusion}
We introduce an entity-centric approach to IR that finds relevant evidence required to answer multi-hop questions from a corpus containing millions of paragraphs leading to significant improvement to an existing QA system. 


\section*{Acknowledgements}
This work is funded in part by the Center for Data Science and the Center for Intelligent Information
Retrieval, and in part by the National Science Foundation under Grant No. IIS-1514053 and in part
by the International Business Machines Corporation Cognitive Horizons Network agreement number
W1668553 and in part by the Chan Zuckerberg Initiative under the project Scientific Knowledge
Base Construction. Any opinions, findings and conclusions or recommendations expressed in this
material are those of the authors and do not necessarily reflect those of the sponsor.
\bibliography{emnlp-ijcnlp-2019}
\bibliographystyle{acl_natbib}
\end{document}